\documentclass[10pt,conference]{IEEEtran}
\IEEEoverridecommandlockouts
\usepackage{cite}
\usepackage{amsmath,amssymb,amsfonts}
\usepackage{algorithmic}
\usepackage{graphicx}
\usepackage{textcomp}
\usepackage{xcolor}
\usepackage[utf8]{inputenc}
\usepackage{booktabs} 
\usepackage{rotating} 
\usepackage{array}    
\usepackage[english]{babel}
\usepackage{caption} 
\usepackage{subcaption} 
\usepackage{adjustbox} 
\usepackage{hyperref}
\usepackage{newunicodechar}
\newunicodechar{₂}{$_2$}
\usepackage{placeins}
\usepackage{makecell}
\usepackage{tabularx}
\usepackage{multirow}
\usepackage{siunitx}
\usepackage{float} 
\usepackage{stfloats}

\def\BibTeX{{\rm B\kern-.05em{\sc i\kern-.025em b}\kern-.08em
    T\kern-.1667em\lower.7ex\hbox{E}\kern-.125emX}}

\begin{document}

\title{Robust gene prioritization for Dietary Restriction via Fast-mRMR Feature Selection techniques}



\author{Rubén Fernández-Farelo$^1$, Jorge Paz-Ruza$^1$, Bertha Guijarro-Berdiñas$^1$, Amparo Alonso-Betanzos$^1$, and Alex A. Freitas$^2$ \\

\{ruben.fernandez.farelo, j.ruza, berta.guijarro, amparo.alonso.betanzos\}@udc.es, A.A.Freitas@kent.ac.uk

\thanks{j.ruza@udc.es, This work is funded by MICIU/AEI/ 10.13039/501100011033 together with ERDF/EU (Grant PID2023-147404OB-I00) and with ESF+ (FPU21/05783), Ministry for Digital Transformation and Civil Service and `Next-GenerationEU'/PRTR (TSI-100925-2023-1), and Xunta de Galicia (ED431C 2022/44, ED431G 2023/01).}

\vspace{.3cm}\\

1 - Universidade da Coruña, Grupo LIDIA - CITIC, A Coruña 15071, Spain \\
2 - School of Computing, University of Kent, Canterbury CT2 7FS, United Kingdom

}

\maketitle

\begin{abstract}
 Gene prioritization (identifying genes potentially associated with a biological process) is increasingly tackled with Artificial Intelligence. However, existing methods struggle with the high dimensionality and incomplete labelling of biomedical data. This work proposes a more robust and efficient pipeline that leverages Fast-mRMR Feature Selection to retain only relevant, non-redundant features for classifiers, building simpler, more interpretable and more efficient models. Experiments in our domain of interest, prioritizing genes related to Dietary Restriction (DR), show significant improvements over existing methods and enable us to integrate heterogeneous biological feature sets for better performance, a strategy that previously degraded performance due to noise accumulation. This work focuses on DR given the availability of curated data and expert knowledge for validation, yet this pipeline would be applicable to other biological processes, proving that feature selection is critical for reliable gene prioritization in high-dimensional omics.

\end{abstract}


\section{Introduction}
Gene prioritization, the task of identifying genes relevant to specific biological mechanisms, is crucial for advancing our understanding of complex processes like Dietary Restriction \cite{fontana2015promoting} (DR, i.e., decreasing nutrient intake without malnutrition), an anti-aging intervention. From a Machine Learning perspective, gene prioritization faces two main challenges: 1) Positive-unlabelled data, with presence of unlabelled examples (genes with no verified relation to the process of interest) and absence of negative examples \cite{bekker2020learning}, which was tackled by recent efforts \cite{vega2022machine, PazRuza2024}, and 2) the high dimensionality of feature spaces, a problem common to biomedical data.

This work fills this gap by implementing a robust Feature Selection pipeline for prioritization of DR genes. While FS is a standard practice, its integration in this task is novel: 
existing baselines \cite{vega2022machine, PazRuza2024} assume relevance of all curated biological features, ignoring the redundancy and uncertainty of biomedical data. We challenge this assumption by integrating the Fast-mRMR algorithm \cite{RamirezGallego2017}, an ideal fit for genomic data by balancing feature relevance with redundancy reduction \cite{peng2005feature} while remaining more efficient for high-dimensional omics data than deep learning-based or recursive FS methods.

Furthermore, to address the computational bottlenecks of high-dimensional biological feature sets (e.g. coexpression data), we introduce task-specific adaptations (quantization and Feature Bagging). Our computational experiments show our FS-aided gene prioritization improves predictive performance and builds more efficient models, both in feature sets with curated or raw biological knowledge; consequently, we use our trained models to produce a ranking of promising candidate genes potentially related to DR.

\paragraph*{Scope of this work} Our proposed FS-based pipeline for gene prioritization is domain-agnostic and applicable to other prioritization tasks (e.g., cancer or diabetes). This study focuses exclusively on Dietary Restriction for two reasons: 1) the availability of ML benchmarks established in the literature \cite{vega2022machine}, and 2) the domain expertise of the authors to biologically interpret and validate the top-ranked novel candidates. Extending this validation to other biological tasks without specialized biological supervision would risk generating statistically significant but biologically meaningless results.

\section{Proposed Method}

This section formally introduces the broader gene prioritization task, our of FS-based prioritization pipeline in DR tasks, and the adaptation of the method for massive feature sets of raw genomic data through Bagged FS.

\subsection{Problem Formalization and Related Work}

Let $\mathcal{G} = \{g_1, \dots, g_n\}$, $\mathbf{g}_i \in \mathbb{R}^d$ be the set of candidate genes, each represented by a feature vector of heterogeneous omics data. Gene prioritization tasks handle Positive-Unlabelled (PU) data, where ground truth is partially observed: we possess a subset of experimentally validated positive examples $\mathcal{P} \subset \mathcal{G}$ (here, DR-related genes), while the remaining examples are unlabelled $\mathcal{U} = \mathcal{G} \setminus \mathcal{P}$ (genes with no current experimental evidence of association); explicit negative examples (genes with proved absence of non-association to DR) do not exist. In the particular case of gene prioritization, this evolves to the goal of learning a scoring function $f: \mathbb{R}^d \rightarrow \mathbb{R} \in [0,1]$ that ranks the genes in $\mathcal{U}$, assigning higher probabilities to positive candidates hidden within the unlabeled background.

DR gene prioritization was initially tackled by Vega Magdaleno et al. \cite{vega2022machine} using standard ML techniques under the strong assumption that all unlabelled genes were negative examples. Later, Paz-Ruza et al. \cite{PazRuza2024} addressed the PU nature of the task through PU Learning techniques. Our work directly builds upon the robust task formulations of these previous works, but explicitly addresses through Feature Selection the critical challenge of high dimensionality, uncertainty and redundancy of genomic feature sets, which previous approaches neglected.

\begin{figure*}[!t]
    \centering
    
    \begin{subfigure}[b]{0.9\textwidth}
        \centering
    \includegraphics[width=0.85\linewidth]{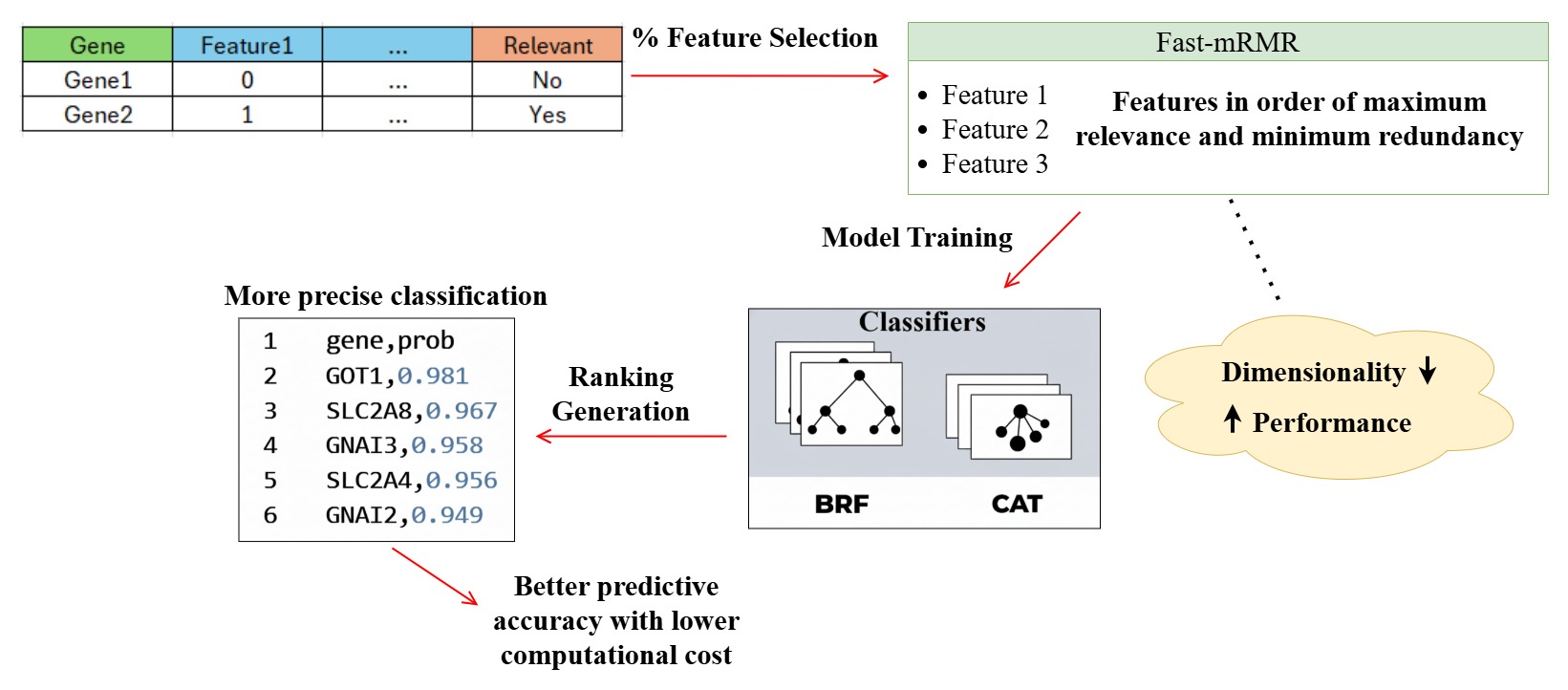}
        \caption{\textbf{Pipeline Architecture.} Overview of the proposed approach for robust gene prioritization.}
        \label{fig:pipeline_main}
    \end{subfigure}
    
    \vspace{0.3cm} 
    
    \begin{subfigure}[b]{0.55\textwidth} 
        \centering
        \includegraphics[width=\linewidth]{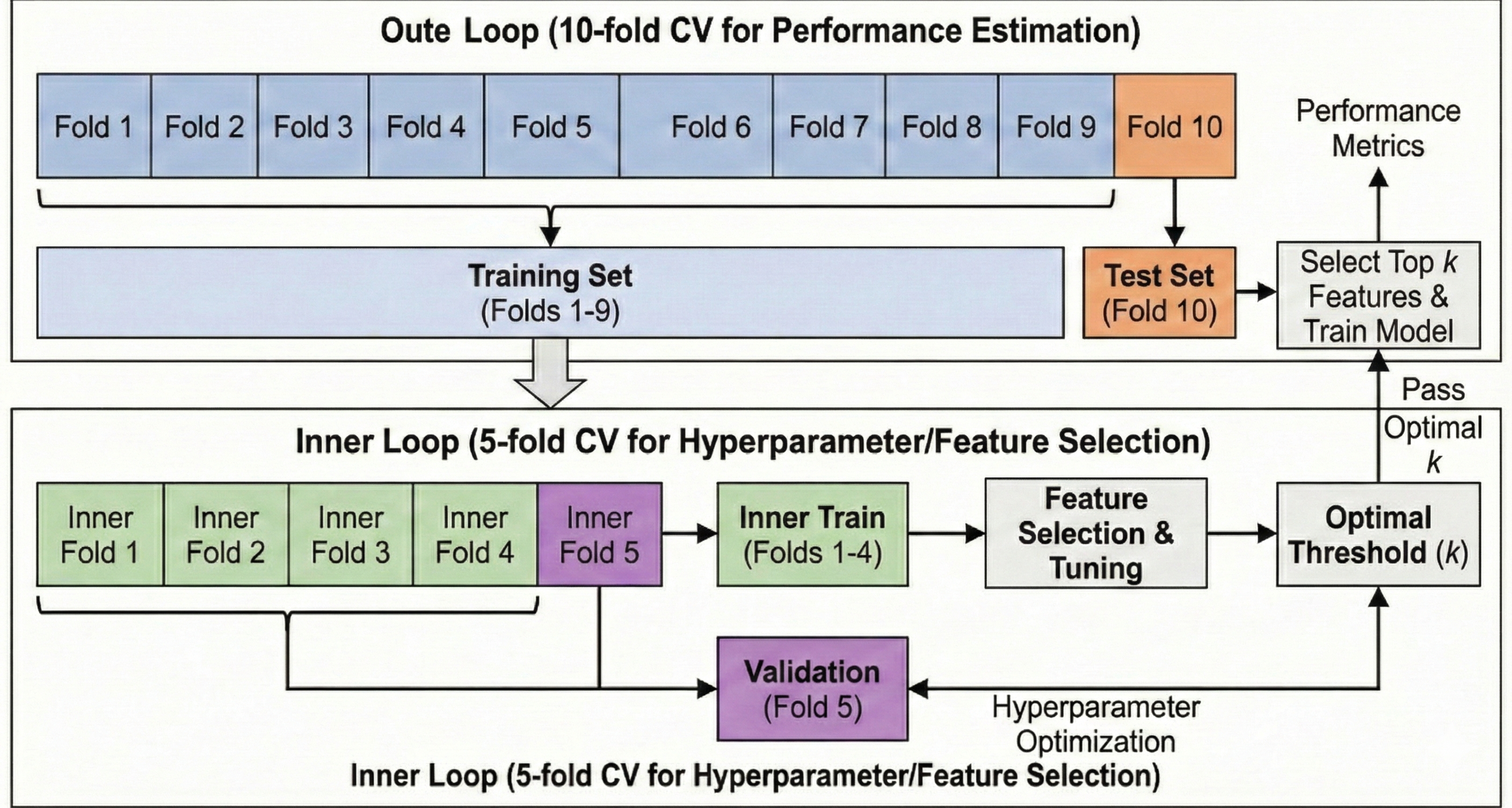}
        \caption{\textbf{Training Block.} Detail of the Nested Cross-Validation strategy.}
        \label{fig:training_detail}
    \end{subfigure}
    \hfill 
    \begin{subfigure}[b]{0.40\textwidth} 
        \centering
        \includegraphics[width=\linewidth]{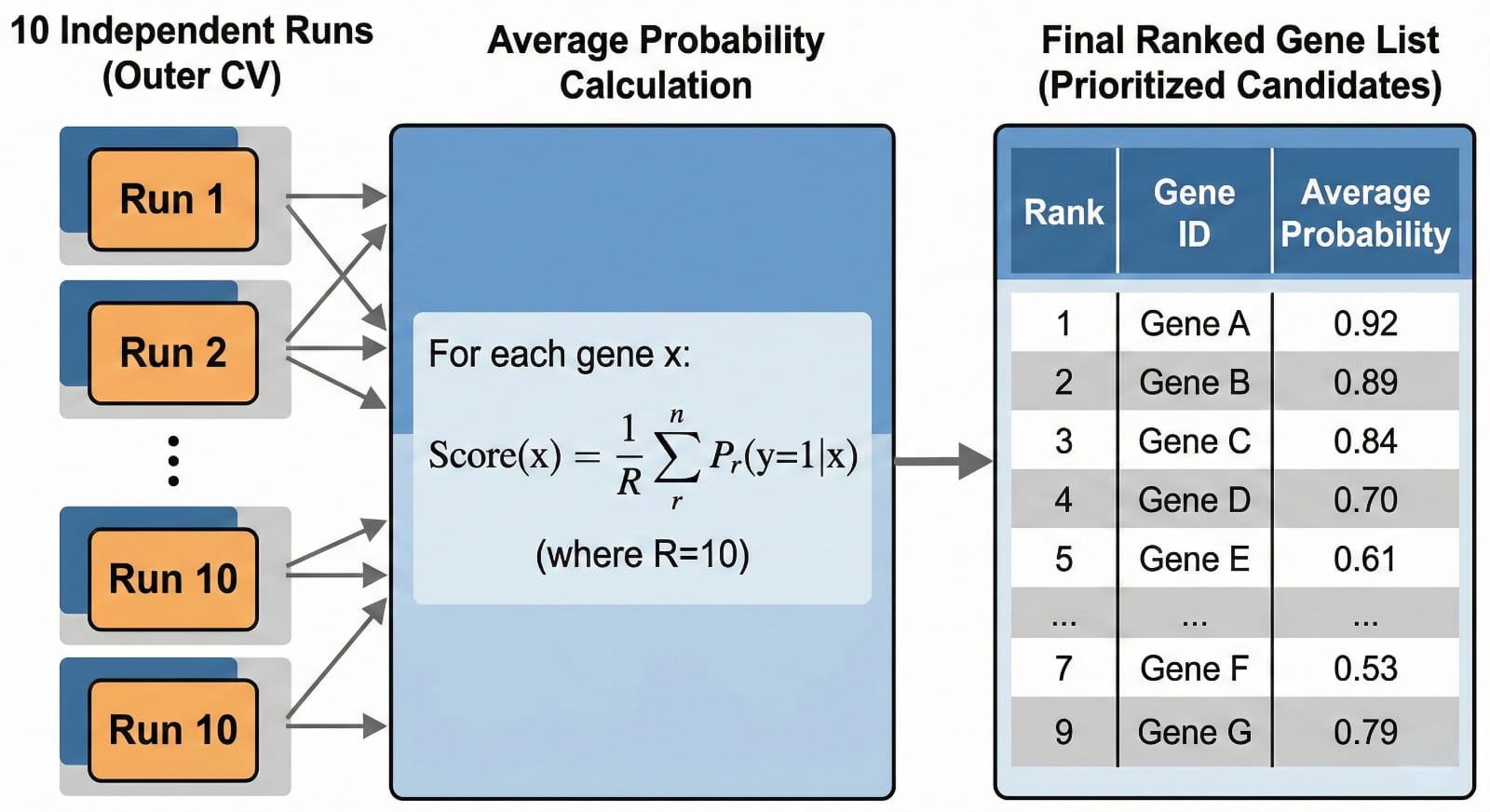}
        \caption{\textbf{Prediction Block.} Detail of the ranking generation.}
        \label{fig:ranking_detail}
    \end{subfigure}
    
    \caption{The proposed prioritization framework: (a) pipeline overview, (b) nested CV training, and (c) ranking generation.}
    \label{fig:combined_architecture}
\end{figure*}

\subsection{The Pipeline Architecture}
Our proposed gene prioritization pipeline, depicted in Figure \ref{fig:combined_architecture}, addresses the dual challenge of high dimensionality ($d \gg n$) and label uncertainty, and is structured into three key phases: 

\subsubsection{Feature Selection in a PU Context}
First, we apply the Fast-mRMR \cite{RamirezGallego2017} algorithm to select features that maximize relevance and minimize inter-feature redundancy. This algorithm was chosen due to its proven suitability for omics data and its computational efficiency over other classic and modern approaches, which is critical for our use-case \cite{RamirezGallego2017}. With this, we reduce the feature set into a compact and informative subset without the prohibitive costs of wrapper or iterative methods.

To address the absence of negative examples, required to compute mutual information in Fast-mRMR, we adopt standard PU Learning assumptions of separability and uniformness during the FS phase  \cite{bekker2020learning}: unlabeled examples (i.e. genes) are treated as provisional negative examples against the known positives. This is possible because filter methods like mRMR are sufficiently robust to identify features that characterize the minority class $\mathcal{P}$ against the general genomic background \cite{RamirezGallego2017}.

\subsubsection{Model Training}
Following standard methodology, model training follows a Nested Cross-Validation (CV) scheme (Figure \ref{fig:combined_architecture}(b)). We consider ensemble classifiers to be independently trained and evaluated in a dual-loop structure:

\begin{itemize}
    \item The Inner CV (5-fold) optimizes the percentage $k$ of features to be selected by FastMRMR in the outer loop.
    \item The Outer CV (10-fold) selects (using each train fold) the most relevant genomic features and trains the ensemble classifiers; the model is evaluated on each test fold using such selected features, and scores each gene for ranking.
\end{itemize}

Fast-mRMR is re-executed within each fold to avoid FS-driven data leakage, which is common in ML omics research. 

\subsubsection{Ranking Generation}
Despite the subrogation to a binary classification task, gene prioritization still requires a continuous score to order candidates. To ensure prediction stability, a final consensus ranking is obtained by aggregating predicted probabilities across $R=10$ independent runs (Figure \ref{fig:combined_architecture}(c)), calculating the final score for a candidate gene $g_i$ as the ensemble average of its predicted probabilities whenever it appeared in the test fold of each run:
\begin{equation}
    Score(g_i) = \frac{1}{R} \sum_{r=1}^{R} P_{r}(y=1|g_i)
\end{equation}

\subsection{Scalability Adaptation: Feature Bagging Strategy}
\label{sec:bagging_strategy}

\begin{figure}[htbp]
    \centering
    \includegraphics[width=0.80\columnwidth]{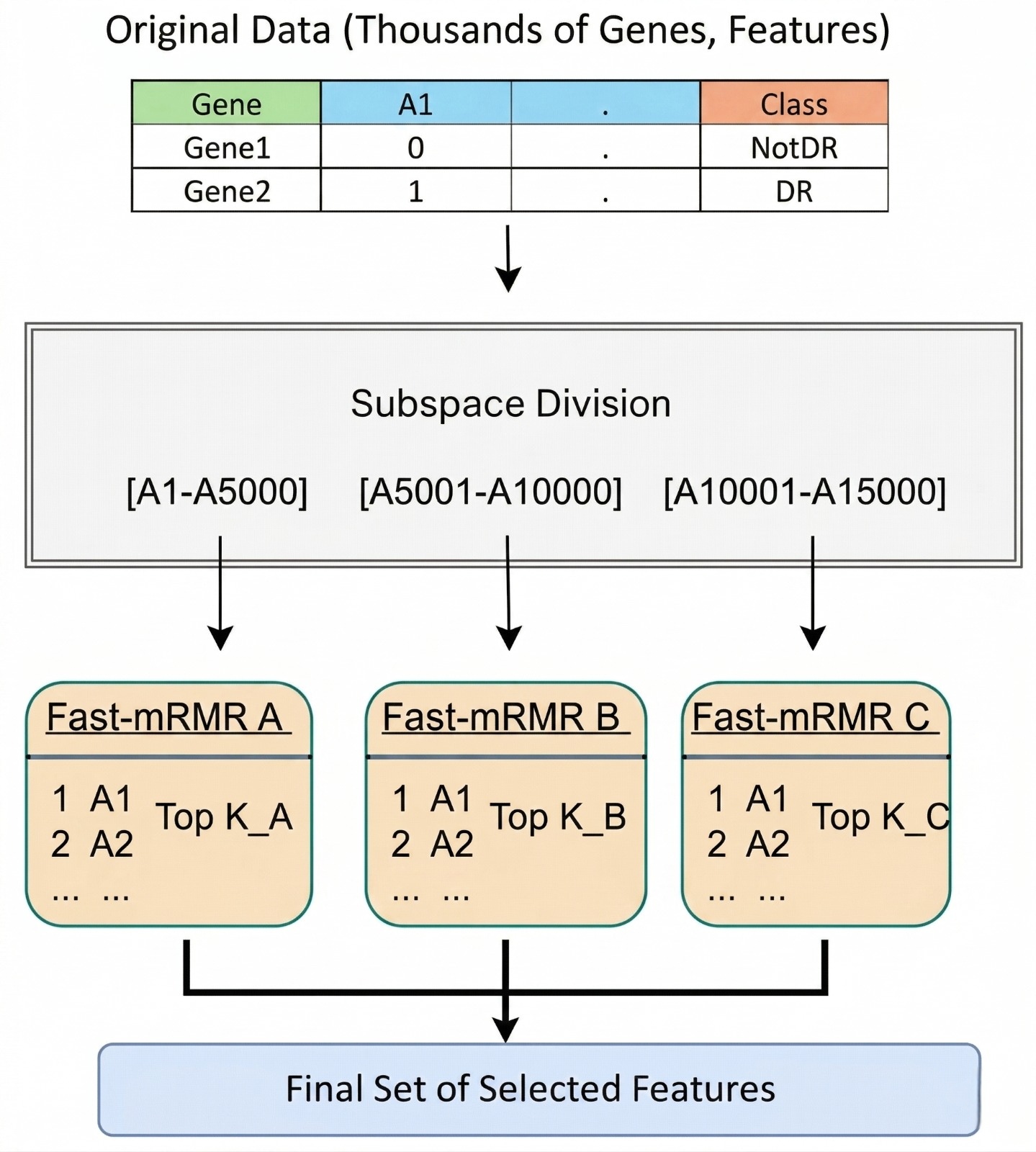} 
    \caption{Feature Bagging strategy used to handle high-dimensional raw genomic data.}
    \label{fig:feature_bagging}
\end{figure}

To handle high-dimensional raw genomic data (e.g. gene coexpressions), where directly applying FS is computationally prohibitive, we implemented a Feature Bagging strategy. The feature space $A = \{A_1, A_2, \dots, A_n\}$ was split into disjoint blocks (e.g., $A_1-A_{5000}$, as shown in Fig. \ref{fig:feature_bagging}); Fast-mRMR is then applied per-block to obtain its top $K$ most relevant features (Top $K\_A$, Top $K\_B$ \ldots), and the partial subsets are finally aggregated to form the final reduced training set.

\section{Experimental Setup}

This section contains information on the specific feature sets, classifiers and implementation details of our experiments.

\subsection{Feature sets and Models} 

Table \ref{tab:datasets} describes the feature sets employed in our experimentation: Gene Ontology (GO) \cite{ashburner2000gene} and PathDIP \cite{rahmati2017pathdip}. These feature sets originate from Vega Magdaleno et al.'s work \cite{vega2022machine}, where readers can find more details on the dataset construction and biological labeling process. We also employ the genes' coexpression data to test scalability through the Bagged FS adaptation, as well as the combination of GO and PathDIP to test our method's ability to integrate heterogeneous feature sets. For numerical stability, continuous features were discretized using quantile binning with 10 intervals.

Regarding models, we follow the rationale of previous works \cite{vega2022machine, PazRuza2024} and employ boosting (CatBoost, CAT) and bagging (Balanced Random Forest, BRF) ensembles, which remain the state-of-the-art for structured tabular data, even over deep learning methods \cite{Grinsztajn2022}. Furthermore, the intrinsic explainability of ensembles is critical in gene prioritization tasks to ensure that the identified candidates can be biologically interpreted and validated by domain experts \cite{Lundberg2020, PazRuza2024}.



\subsection{Implementation and Hyperparameters} 
To ensure a fair and direct comparison with previous benchmarks, all classifier hyperparameters for Balanced Random Forest (BRF) and CatBoost (CAT) were kept identical to those established in \cite{vega2022machine, PazRuza2024}, including the use of 500 estimators for both ensembles. The inner Cross-Validation (CV) loops were employed to optimize the feature selection threshold within a range of 2.5\% to 60\% of the total features. These boundaries were determined through preliminary experiments, which showed that thresholds below 2.5\% led to significant information loss and performance degradation. Final thresholds were selected based on these inner loop validations to achieve an optimal balance between feature reduction and predictive accuracy. All experiments were conducted on a workstation with an Intel Core i7-11700 (8 cores) and 16 GB RAM. The code is available for reproducibility.\footnote{\url{https://github.com/ru-farelo/fs-dr-gene-prioritization}}


\begin{table}[htbp]
\centering
\caption{Statistics for the experimental feature sets.}
\label{tab:datasets}
\resizebox{\columnwidth}{!}{%
\begin{tabular}{lrrrrr}
\toprule
\textbf{Feature Set} & \textbf{Features} &
\makecell{\textbf{Total}\\\textbf{Genes}} &
\makecell{\textbf{Total}\\\textbf{Positives}} &
\makecell{\textbf{Common}\\\textbf{Genes}} &
\makecell{\textbf{Common}\\\textbf{Positives}} \\ 
\midrule
GO & 8,640 & 1,124 & 114 & 981 & 109 \\
PathDIP & 1,640 & 986 & 110 & 981 & 109 \\
Coexpression & 44,946 & 1,048 & 113 & - & - \\
\bottomrule
\end{tabular}%
}
\end{table}

\section{Results}

This section contains the results of the computational experiments in individual, combined and large-scale feature sets for DR gene prioritization, as well as a biologically-grounded analysis of the most relevant features and most promising candidate genes obtained by the classifiers.

\subsection{Computational Experiments}

Table \ref{tab:full_comparison} presents the performance comparison of our proposed pipeline against the Original and PU Learning baselines. We report results for both CAT and BRF classifiers across all configurations. The analysis yields two major findings:

\begin{table}[ht]
\centering
\caption{Performance comparison. Reported values represent the mean score averaged over 10 independent experimental runs. Best results per metric are bolded; among those, $\dagger$ denotes statistical significance ($p<0.05$) compared to all Original and PU Learning baselines. An asterisk (*) denotes an improvement on a given metric for a feature set and model combination compared to the Original and PU baselines.}
\label{tab:full_comparison}
\resizebox{\columnwidth}{!}{%
\begin{tabular}{llrrr}
\toprule
\textbf{Method} & \textbf{Configuration} & \textbf{AUC-ROC} & \textbf{G-Mean} & \textbf{F1-Score} \\
\midrule
\multirow{4}{*}{\textbf{\makecell[l]{Original\\(No Selection)}}} 
 & PathDIP / CAT & 0.813 & 0.702 & 0.505 \\
 & PathDIP / BRF & 0.827 & 0.771 & 0.476 \\
 & GO / CAT      & 0.838 & 0.684 & 0.498 \\
 & GO / BRF      & 0.835 & 0.763 & 0.389 \\
\addlinespace 
\multirow{4}{*}{\textbf{PU Learning}}
 & PathDIP / CAT & 0.829 & 0.749 & 0.531 \\
 & PathDIP / BRF & 0.827 & 0.768 & 0.461 \\
 & GO / CAT      & 0.839 & 0.729 & 0.496 \\
 & GO / BRF      & 0.829 & 0.762 & 0.380 \\
\midrule
\multirow{6}{*}{\textbf{\makecell[l]{Fast-mRMR}}}
 & PathDIP / CAT & 0.830* & 0.732 & 0.523 \\
 & PathDIP / BRF & 0.841* & \textbf{0.779}* $\dagger$ & 0.479* \\ 
 & GO / CAT      & \textbf{0.852}* $\dagger$ & 0.738* & \textbf{0.532}* \\ 
 & GO / BRF      & 0.842* & 0.766* & 0.450* \\
 \addlinespace
 & GO+PathDIP / CAT & \textbf{0.873}* $\dagger$ & 0.750* & \textbf{0.560}* \\
 & GO+PathDIP / BRF & 0.869* & \textbf{0.798}* $\dagger$ & 0.503* \\
\bottomrule
\end{tabular}%
}
\end{table}

\begin{itemize}
    \item \textbf{Superiority of Feature Selection:} The Fast-mRMR method significantly outperforms the baselines even in single-source configurations. Specifically, it achieves statistically significant improvements ($\dagger$) in AUC-ROC for GO (0.852) and G-Mean (Geometric Mean of Sensitivity and Specificity) for PathDIP (0.779) compared to the Original and PU Learning strategies.
    
    \item \textbf{Effective Data Integration:} Most notably, the GO+PathDIP (Combined) configuration achieves the best overall performance, with CatBoost reaching the highest AUC-ROC (0.873) and F1-Score (0.560). This outcome reverses the conclusions from previous works \cite{vega2022machine}, where combining these datasets without feature selection degraded performance due to the curse of dimensionality. Our results confirm that robust FS is the key mechanism that unlocks the synergy between heterogeneous biological data sources.

    \item \textbf{Feature Selection Stability:} To ensure the reliability of the prioritized genes, we evaluated the stability of Fast-mRMR using the Jaccard Index (Intersection over Union) across the experimental folds. The pipeline exhibited high consistency with an average feature overlap of $71.1\% \pm 3.5\%$. The highest stability was observed in the PathDIP dataset ($75.9\% \pm 3.3\%$), while GO and the combined GO+PathDIP sets maintained robust overlaps of $68.1\%$ and $67.5\%$, respectively. The high Jaccard scores confirm the robustness of the Fast-mRMR selection process, providing a stable foundation for the subsequent prioritization of novel candidates.
\end{itemize}

\subsubsection{Influence of Feature Selection Thresholds}
\label{sec:threshold_influence}

\begin{figure}[h]
    \centering
    \begin{subfigure}[b]{0.24\textwidth}
        \centering
        \includegraphics[width=\linewidth]{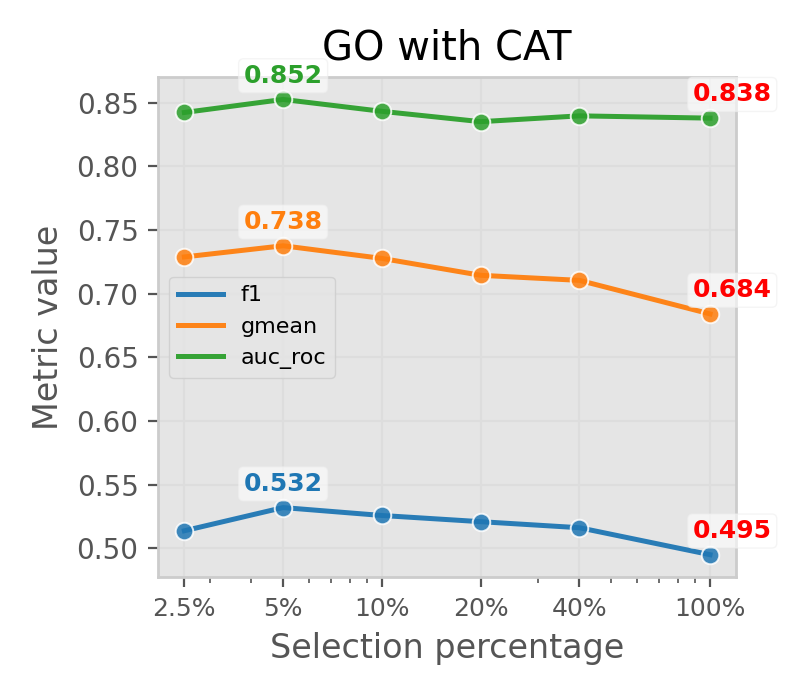} 
        \label{fig:go_cat}
    \end{subfigure}%
    \hfill
    \begin{subfigure}[b]{0.24\textwidth}
        \centering
        \includegraphics[width=\linewidth]{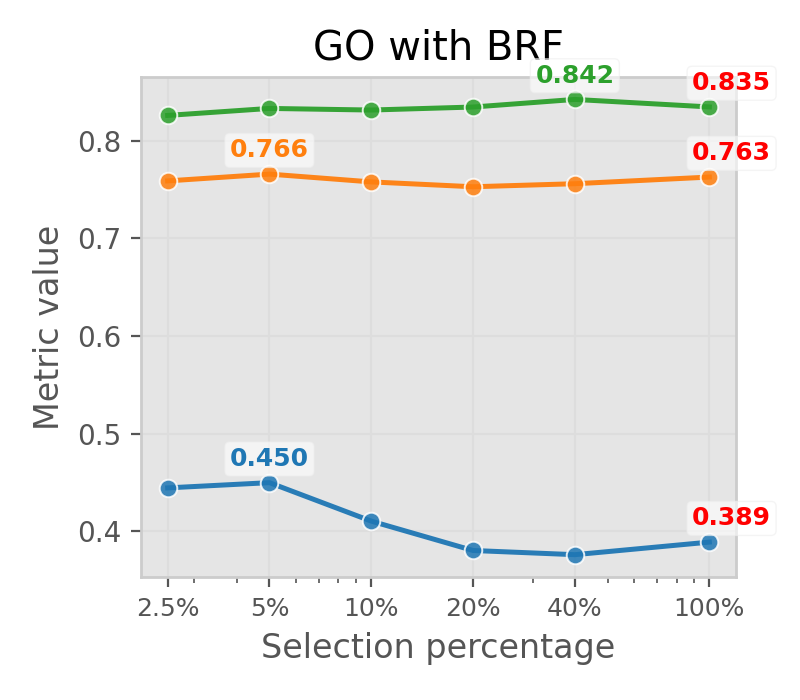}
        \label{fig:go_brf}
    \end{subfigure}%
    \hfill
    \begin{subfigure}[b]{0.24\textwidth}
        \centering
        \includegraphics[width=\linewidth]{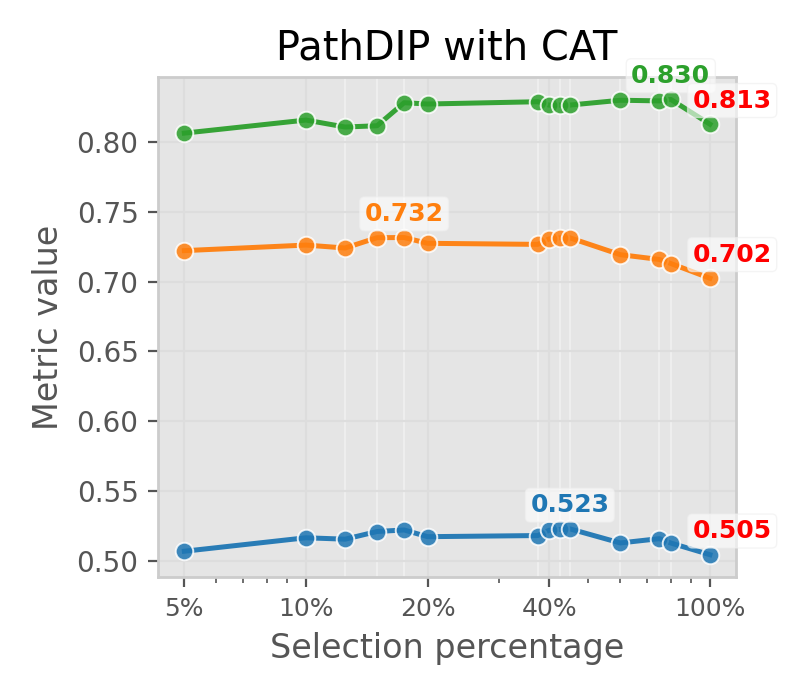}
        \label{fig:path_cat}
    \end{subfigure}%
    \hfill
    \begin{subfigure}[b]{0.24\textwidth}
        \centering
        \includegraphics[width=\linewidth]{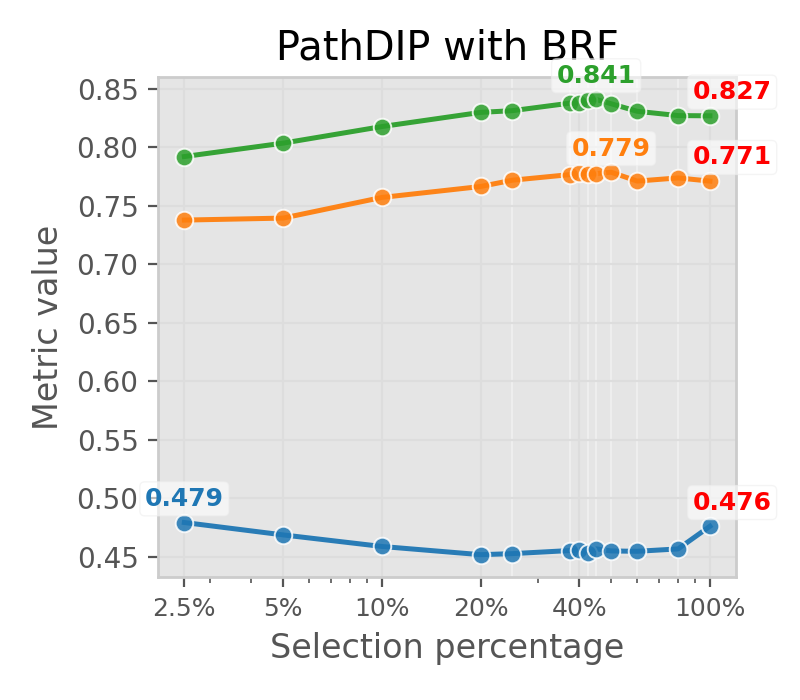}
        \label{fig:path_brf}
    \end{subfigure}
    
    \caption{Performance evolution (AUC-ROC, F1, G-Mean) on single-source datasets as a function of the percentage of features selected by Fast-mRMR.}
    \label{fig:four_plots}
\end{figure}

To quantify the impact of feature selection on model stability, we analyzed the evolution of predictive metrics across varying thresholds ($k$) on the single-source datasets. Figure \ref{fig:four_plots} displays the performance trajectories for CatBoost and BRF from 2.5\% to 100\% feature retention. The results reveal a consistent trend: predictive power is maximized at low selection ratios across all configurations. The high-dimensional GO dataset peaks at just 5\% of features, while PathDIP stabilizes near 25\%. Crucially, increasing the feature set beyond these points yields diminishing returns or performance degradation. This confirms that the vast majority of the original omics features are redundant and constitute biological noise, which the Fast-mRMR stage effectively filters out to enhance model discriminability.

\subsection{Scalability on High-Dimensional Data}

We evaluated the scalability of the pipeline on the Coexpression dataset using the Feature Bagging strategy. The results in Table \ref{tab:coexpression_results} confirm that the pipeline successfully handles high dimensionality. While the baseline models exhibited performance close to random guessing (AUC-ROC $\approx 0.50$), the proposed Fast-mRMR strategy consistently improved performance over the Original method (marked with * in Table \ref{tab:coexpression_results}). Specifically, BRF achieved improvements across all metrics, while CatBoost reached the highest overall AUC-ROC (0.547). While these absolute values appear modest, they must be interpreted within the strict PU evaluation framework, where penalizing latent positive discoveries as False Positives mathematically underestimates all standard performance metrics (AUC-ROC, F1-Score and G-Mean). Hence, these scores cannot be taken as true absolute values, but rather as conservative lower bounds of the actual performance. Crucially, these gains were achieved while retaining less than 5\% of the original feature space (fewer than 2,500 features out of $\approx$45,000), proving that the method can extract meaningful biological signal and make the classification task computationally manageable.

\begin{table}[h]
\centering
\caption{Performance comparison on the Coexpression dataset ($>44,000$ features). * indicates an improvement over non-FS methods. Best results per metric are bolded, and values in parenthesis represent the percentage of selected features.}
\label{tab:coexpression_results}
\resizebox{\columnwidth}{!}{%
\begin{tabular}{ll ccc}
\toprule
\textbf{Method} & \textbf{Classifier} & \textbf{AUC-ROC} & \textbf{G-Mean} & \textbf{F1-Score} \\
\midrule
\multirow{2}{*}{\makecell[l]{Original\\(No Selection)}} 
 & BRF & 0.504 & 0.502 & 0.179 \\
 & CatBoost & 0.510 & 0.070 & 0.038 \\
\midrule
\multirow{2}{*}{\makecell[l]{\textbf{Fast-mRMR}\\(Proposed)}} 
 & BRF & \textbf{0.542*} (2.5\%) & \textbf{0.512}* (2.5\%) & \textbf{0.185}*  (2.5\%) \\
 \addlinespace
 & CatBoost & \textbf{0.547}*  (5\%) & 0.065  (10\%) & 0.035  (10\%) \\
\bottomrule
\end{tabular}%
}
\end{table}


\subsection{Computational Efficiency and Sustainability Analysis}

\begin{figure}[htbp!]
    \centering
    \includegraphics[width=0.9\linewidth]{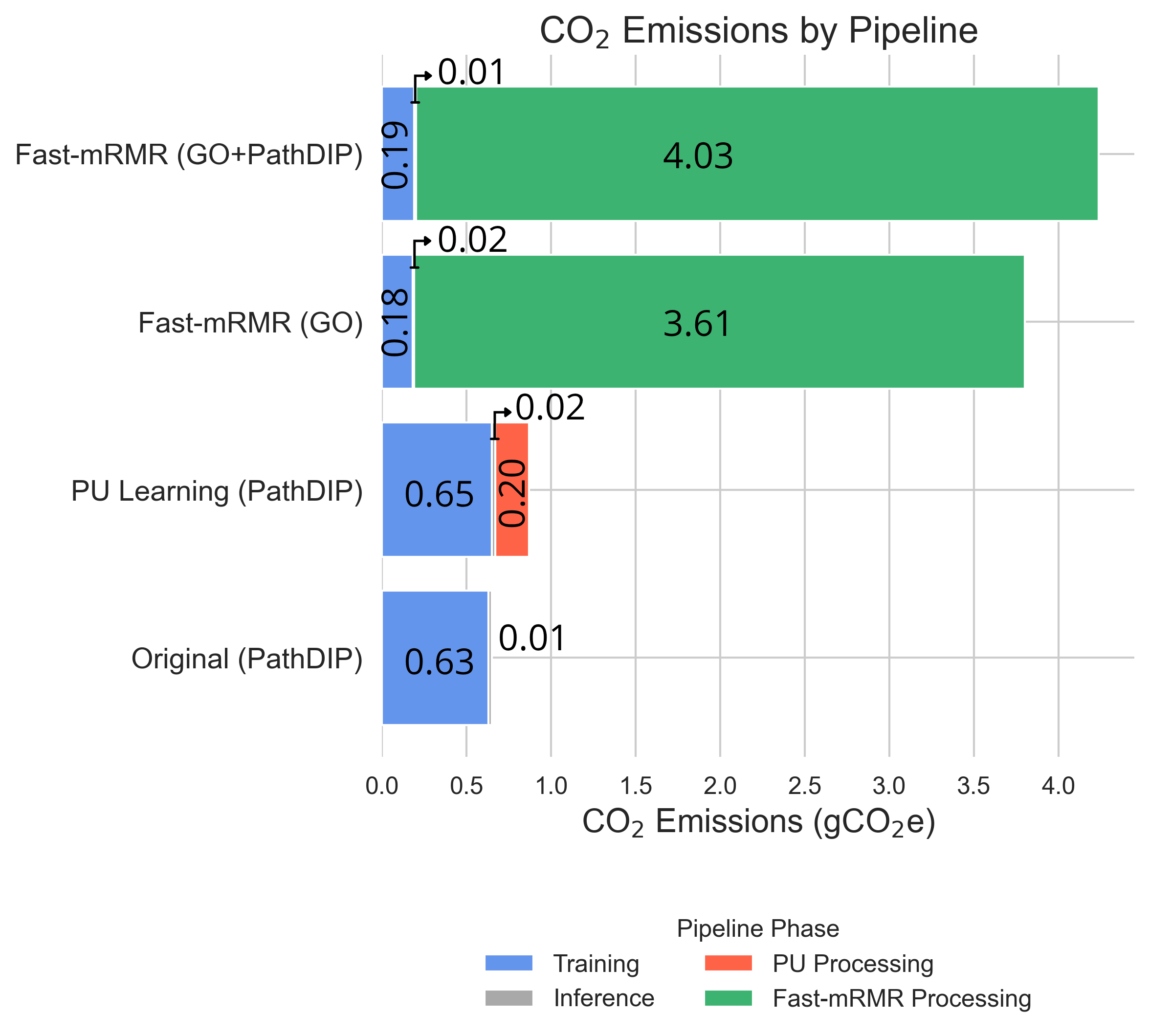}
\hfill
\vspace{0.2cm}
    \includegraphics[width=0.9\linewidth]{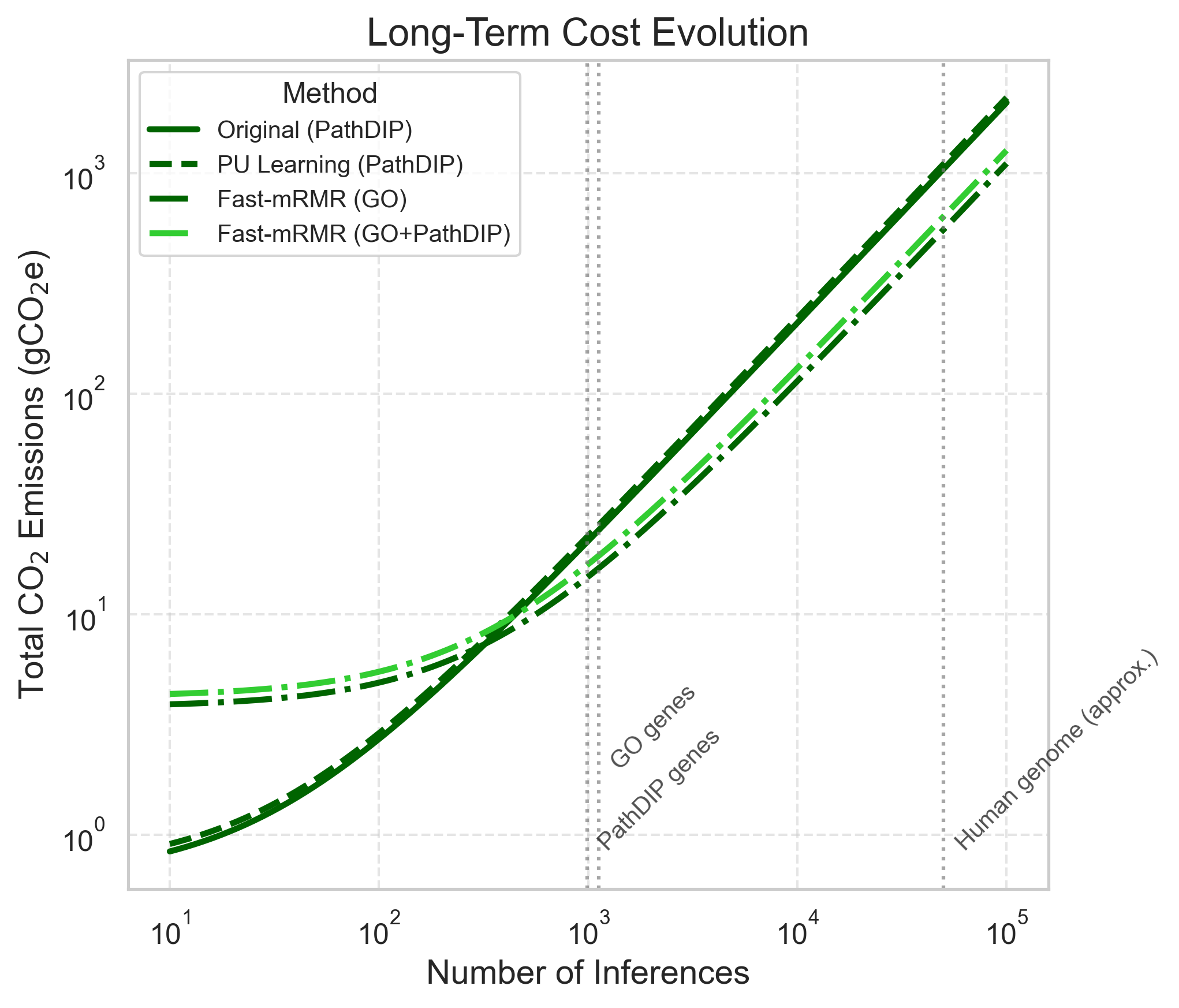}
\caption{Computational efficiency per phase (top) and long-term evolution of prioritization efficiency (bottom).}
\label{fig:efficiency_analysis}
\end{figure}

We analyzed computational efficiency via CO₂ emissions using \textit{CodeCarbon} (Figure \ref{fig:efficiency_analysis}). For a fair comparison, we normalized the PU\_Processing cost, considering the computational cost of a single run of the 10-fold cross-validation. 

Our Fast-mRMR approach represents a strategic trade-off: while it introduces an initial Feature Selection phase, it drastically reduces the subsequent Training cost, which is the main bottleneck in the baseline methods. As shown in the bottom panel of Fig. \ref{fig:efficiency_analysis}, this initial investment results in a significantly lower per-inference cost (a flatter slope). This confirms that our approach is a greener and more scalable solution for long-term use, where the upfront cost is quickly amortized over thousands of predictions. Fast-mRMR(PathDIP) offered no advantage in this regard and was omitted.

Beyond performance, this efficiency supports AI democratization in bioinformatics. Lower computational demands allow small labs with limited hardware to run complex genomic analyses, promoting broader adoption and reproducibility. Moreover, this reduced energy consumption aligns with Ethical Guidelines for Trustworthy AI, particularly environmental well-being. By proposing a Green AI model that lowers carbon footprint without sacrificing accuracy, this work supports more responsible and sustainable research.

\begin{table*}[!t]
\centering
\caption{Top 5 most important features for predicting DR-relatedness with the Original \cite{vega2022machine} and PU Learning \cite{PazRuza2024} methods (both using PathDIP features), and our proposed Fast-mRMR methods: (1) using GO features and (2) using GO+PathDIP features. Feature importance was calculated following the procedure in \cite{PazRuza2024}. Feature definitions are provided in the legend below.}
\label{tab:feature_comparison_full}
\begin{adjustbox}{width=\textwidth}
\begin{tabular}{lr @{\hskip 1.5em} lr @{\hskip 1.5em} lr @{\hskip 1.5em} lr}
\toprule
\multicolumn{2}{c}{\textbf{Original Method}} & 
\multicolumn{2}{c}{\textbf{PU Learning Method}} & 
\multicolumn{2}{c}{\textbf{Proposed Fast-mRMR (1)}} & 
\multicolumn{2}{c}{\textbf{Proposed Fast-mRMR (2)}} \\
\cmidrule(r){1-2} \cmidrule(lr){3-4} \cmidrule(lr){5-6} \cmidrule(l){7-8}
\textbf{Feature} & \textbf{Score} & 
\textbf{Feature} & \textbf{Score} & 
\textbf{Feature} & \textbf{Score} & 
\textbf{Feature} & \textbf{Score} \\
\midrule
KEGG.2 & 100.00 & WikiPathways.37 & 100.00 & GO:0007188 & 100.00 & GO:0051130 & 100.00 \\
KEGG.30 & 45.35 & KEGG.2 & 72.52 & GO:0051130 & 95.36 & EHMN.11 & 91.65 \\
NetPath.23 & 38.95 & KEGG.30 & 56.32 & GO:0055114 & 91.10 & HumanCyc.11 & 86.47 \\
REACTOME.10 & 37.79 & WikiPathways.57 & 47.85 & GO:0007187 & 80.38 & WikiPathways.37 & 74.13 \\
WikiPathways.37 & 34.88 & EHMN.11 & 43.97 & GO:0006259 & 65.93 & ACSN2.11 & 70.92 \\
\bottomrule
\end{tabular}
\end{adjustbox}
\vspace{0.2cm} 

\scriptsize
\begin{tabularx}{\textwidth}{lX @{\hskip 2em} lX}
\toprule
\multicolumn{2}{l}{\textit{\textbf{PathDIP Features}}} & 
\multicolumn{2}{l}{\textit{\textbf{GO Features }}} \\
\cmidrule(r){1-2} \cmidrule(l){3-4}
\textbf{Feature} & \textbf{Definition} & \textbf{Feature} & \textbf{Definition} \\
\midrule
KEGG.2 & Autophagy – animal & GO:0007188 & G protein-coupled receptor signaling (cyclic nucleotide) \\
KEGG.30 & Longevity regulating pathway & GO:0051130 & Process that positively regulates cellular organization \\
NetPath.23 & Brain-derived neurotrophic factor (BDNF) signaling pathway & GO:0055114 & Oxidation–reduction process \\
REACTOME.10 & Cellular responses to external stimuli & GO:0007187 & G protein-coupled receptor signaling (phosphoinositide) \\
WikiPathways.37 & NRF2 pathway & GO:0006259 & DNA metabolic process \\
WikiPathways.57 & Glycolysis and gluconeogenesis & & \\
EHMN.11 & Fructose and mannose metabolism & & \\
HumanCyc.11 & Nicotine degradation IV & & \\
ACSN2.11 & Antioxidant Response & & \\
\bottomrule
\end{tabularx}
\end{table*}

\subsection{Biological analysis of the most relevant features and the most promising novel candidate DR genes} 

Table \ref{tab:feature_comparison_full} shows the 5 top-ranked features in models learned by other methods \cite{vega2022machine, PazRuza2024} and by two versions of our method (selecting GO features and GO+PathDIP features). Two of the top-5 PathDIP features found by our method, EHMN.11 and WikiPathways.37, were also in the top-5 features found by the methods in \cite{vega2022machine, PazRuza2024}; whilst HumanCyc.11 and ACSN2.11 are new top-ranked PathDIP features found by our method. The most common feature across all methods is WikiPathways.37 (NRF2 pathway), found in all 3 sets of top-ranked PathDIP features. Dietary restriction activates Nrf2 and triggers its protective anti-oxidant effects, which help to prevent age-related neurodegenerative disorders \cite{Vasconcelos2019Nrf2}. 

Table \ref{tab:comparacion_genes} shows the 7 top-ranked new candidate DR-related genes found by methods in \cite{vega2022machine, PazRuza2024} and by two versions of our method (selecting 5\% of GO features or 5\% of GO+PathDIP features). These are promising candidate DR-related genes because they are annotated as negatives (non-DR-related) in the dataset but predicted with the highest probabilities to be positives (DR-related).
Among the top-ranked genes found by the two versions of our method, three (GCLM, GOT1, GOT2) are also in the set of top-ranked genes found in [3] or [4], whilst the others are novel candidate DR-genes found in this current work. Out of those novel genes, we highlight RRAGD, encoding the RagD protein, a GTPase involved in activating the mTORC1 pathway \cite{LamaSherpa2023gene} -- a major pathway in ageing and amino acid sensing (leading to cell growth).


\begin{table*}[!t]
    \centering
\caption{Top 7 novel candidate genes comparing baselines \cite{vega2022machine, PazRuza2024} with our Fast-mRMR setups: (1) GO, and (2) GO+PathDIP. DR-probabilities were calculated as in \cite{PazRuza2024}. Genes common to multiple rankings are in italics.}

    \label{tab:comparacion_genes}
    \begin{adjustbox}{max width=\textwidth,center}
    \begin{tabular}{c lr @{\hskip 1.5em} lr @{\hskip 1.5em} lr @{\hskip 1.5em} lr}
        \toprule
        & \multicolumn{2}{c}{\textbf{Original Method}} 
        & \multicolumn{2}{c}{\textbf{PU Learning Method}} 
        & \multicolumn{2}{c}{\textbf{Proposed Fast-mRMR (1)}} 
        & \multicolumn{2}{c}{\textbf{Proposed Fast-mRMR (2)}} \\
        \cmidrule(lr){2-3} \cmidrule(lr){4-5} \cmidrule(lr){6-7} \cmidrule(l){8-9}
        \textbf{Rank} & \textbf{Gene} & \textbf{DR-Prob} 
                      & \textbf{Gene} & \textbf{DR-Prob} 
                      & \textbf{Gene} & \textbf{DR-Prob}
                      & \textbf{Gene} & \textbf{DR-Prob} \\
        \midrule
    1 & \textit{GOT2} & 0.86 & \textit{TSC1} & 0.97 & NOS3 & 0.91 & \textit{GCLM} & 0.89 \\
    2 & \textit{GOT1} & 0.85 & \textit{GCLM} & 0.94 & GHRHR & 0.89 & ATP6V1H & 0.82 \\
    3 & \textit{TSC1} & 0.85 & IRS1 & 0.93 & HTR1B & 0.87 & GPT & 0.82 \\
    4 & CTH & 0.85 & PRKAB1 & 0.92 & \textit{GOT2} & 0.87 & RRAGD & 0.81 \\
    5 & \textit{GCLM} & 0.82 & PRKAB2 & 0.90 & \textit{SESN3} & 0.86 & ATP6V0A2 & 0.80 \\
    6 & \textit{IRS2} & 0.80 & PRKAG1 & 0.90 & \textit{GNAS} & 0.85 & \textit{GNAS} & 0.80 \\
    7 & SESN2 & 0.80 & \textit{IRS2} & 0.90 & \textit{GOT1} & 0.84 & \textit{SESN3} & 0.80 \\
        \bottomrule
    \end{tabular}
    \end{adjustbox}
\end{table*}

\section{Conclusions and Future Work}

This study integrates Fast-mRMR into gene prioritization pipelines and proves its effectiveness in filtering out biological noise of high-dimensional omics data. We obtain compact, interpretable models with superior prioritization performance and reduced computational costs, aligning with sustainable and democratic Green AI principles. 

Our results in the prioritization of Dietary Restriction genes also show that robust feature selection is the critical enabler that unlocks the synergy between heterogeneous biological data sources, turning a previously detrimental increase in dimensionality into a statistically significant performance gain.

Crucially, while validated on Dietary Restriction as a proof-of-concept, this methodology is domain-agnostic. The findings confirm that reducing redundancy leads to reliable gene prioritization in high-dimensional tasks, as long as domain expertise is available to validate the biological relevance of the results.

Future work will focus on applying this methodology to other biological processes of interest, and the integration of Fast-mRMR with PU Learning to refine negative instance selection and thus further improve prioritization performance.


\bibliographystyle{IEEEtran}
\bibliography{IEEE}

\end{document}